\documentclass[10pt, conference]{IEEEtran}

\usepackage{amsmath,amssymb,amsfonts}
\usepackage{graphicx}
\graphicspath{{/}{fig/}}
\usepackage{textcomp}

\usepackage{amsthm}
\usepackage{xcolor}
\usepackage{float}

\usepackage{booktabs}
\usepackage{cite}

\usepackage{pgfplots}
\pgfplotsset{compat=1.7}
\usepgfplotslibrary{groupplots}

\usepackage{flushend}
\usepackage{hyperref}

\usepackage[style=base]{caption}
\usepackage{subcaption}

\usepackage{tikz}

\usepackage{pifont}
\newcommand{\cmark}{\ding{51}}%

\newlength\figureheight
\newlength\figurewidth
\title{\LARGE \bf
    AutoSOS: Towards Multi-UAV Systems Supporting Maritime Search and Rescue with Lightweight AI and Edge Computing
}

\author{
    \IEEEauthorblockN{
        Jorge Pe\~{n}a Queralta\textsuperscript{1},
        Jenni Raitoharju\textsuperscript{2},
        Tuan Nguyen Gia\textsuperscript{1}, \\
        Nikolaos Passalis\textsuperscript{2},
        Tomi Westerlund\textsuperscript{1} \\[+6pt]
    }
    \IEEEauthorblockA{
        \textsuperscript{1} \href{https://tiers.utu.fi}{Turku Intelligent Embedded and Robotic Systems (TIERS) Lab, University of Turku, Finland} \\
        \textsuperscript{2} Unit of Computing Sciences, Tampere University, Finland \\
        Emails: \textsuperscript{1}\{jopequ, tunggi, tovewe\}@utu.fi, \textsuperscript{2} \{jenni.raitoharju, nikolaos.passalis\}@tuni.fi \\
    }
}

\begin{document}

\maketitle
\thispagestyle{empty}
\pagestyle{empty}

\global\csname @topnum\endcsname 0
\global\csname @botnum\endcsname 0
\begin{abstract}

    Rescue vessels are the main actors in maritime safety and rescue operations. At the same time, aerial drones bring a significant advantage into this scenario. This paper presents the research directions of the AutoSOS project, where we work in the development of an autonomous multi-robot search and rescue assistance platform capable of sensor fusion and object detection in embedded devices using novel lightweight AI models. The platform is meant to perform reconnaissance missions for initial assessment of the environment using novel adaptive deep learning algorithms that efficiently use the available sensors and computational resources on drones and rescue vessel. When drones find potential objects, they will send their sensor data to the vessel to verity the findings with increased accuracy. The actual rescue and treatment operation are left as the responsibility of the rescue personnel. The drones will autonomously reconfigure their spatial distribution to enable multi-hop communication, when a direct connection between a drone transmitting information and the vessel is unavailable.

\end{abstract}

\begin{IEEEkeywords}
    Robotics; Edge Computing; Multi-Robot Systems; Machine Learning; Active perception; Active Vision; Multi-UAV Systems; Formation Control;
\end{IEEEkeywords}

\IEEEpeerreviewmaketitle

\section{Introduction}
\label{sec:intro}

\begin{figure}
    \centering
    \includegraphics[width=0.48\textwidth]{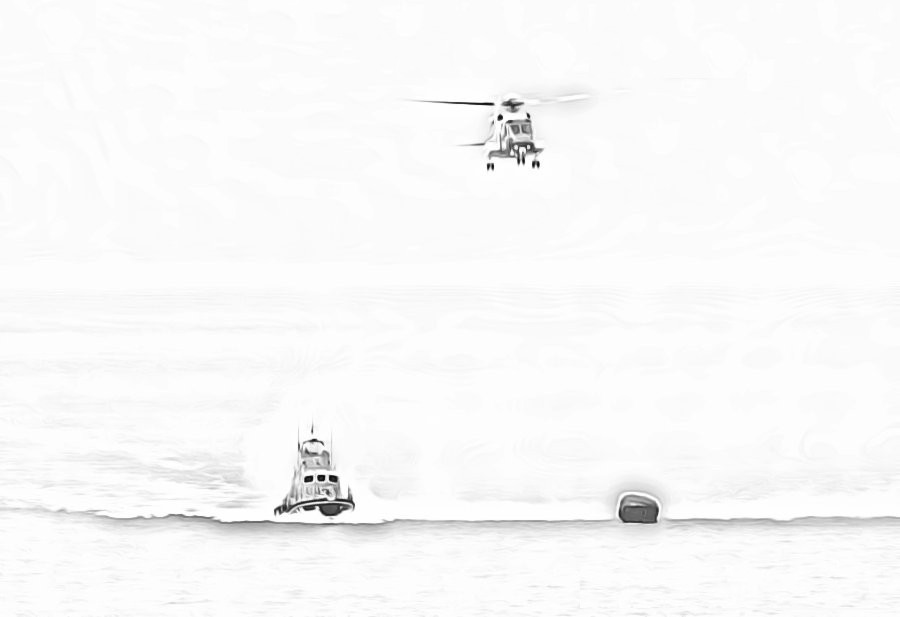}
    \caption{Currently, search and rescue missions in lakes or seas often involve both rescue vessels and helicopters.}
    \label{fig:seasar}
    \vspace{3ex}
\end{figure}

The past decade has seen a boost in the development of autonomous vehicles for civilian use. Google started the development of its self-driving technology for cars in 2009, and since then a myriad of industry leaders, start-ups, and academic researchers have joined the race in the technology sector, a race to make everything autonomous. What started with autonomous cars has now moved into autonomous trucks, autonomous vessels, and autonomous delivery UAVs. In the case of autonomous waterborne transportation, Finland is among the leaders in the race having the world's first fully autonomous commercial ferry in 2018.

Rescue vessels are the main actors in maritime safety and rescue operations. Even though there are fewer obstacles at sea than inland, the constant movement of the sea and waves have a relevant impact on field-of-view from a vessel. UAVs can bring a significant advantage in this scenario. Indeed, UAVs equipped with high-resolution cameras have been used by both police and rescue teams around the globe for search and rescue as well as monitoring purposes. Currently, an aerial view is often provided by helicopters, as illustrated in Fig.~\ref{fig:seasar}. While utilizing multiple helicopters requires extra personnel and significantly increases the cost of the mission, an autonomous multi-UAV system supporting the rescue vessel can provide an strategic advantage and a much higher amount of data in real-time.

Leveraging the state-of-the-art research and innovation in autonomous waterborne navigation, hybrid distributed computing architectures with edge computing, lightweight artificial intelligence models for sensor fusion and object detection in embedded edge devices, and theory of multi-agent systems, we propose a heterogeneous, autonomous multi-robot search and rescue assistance platform. The platform can play a life-saving role in multiple situations, for example, by exploring a predefined area in rivers, lakes, or seas to locate persons or other objects in the water. Only the actual rescue and treatment of the rescued people is left as the responsibility of the rescue personnel. 

The platform can perform reconnaissance missions with an initial assessment of the objective environment using novel adaptive deep learning algorithms that efficiently use the available sensors and computational resources on the UAVs and the boat. When UAVs find potential objects, they will send their sensor data to the vessel, where a more capable computer is able to analyze and verify the findings with increased accuracy. The UAVs will also autonomously reconfigure their spatial distribution to enable multi-hop communication when a direct connection between a UAV transmitting information and the vessel is unavailable. The boat's sensors and computational resources will be exploited and combined with the information from the UAVs.

In this paper, we present the research directions of the AutoSOS project, where we work in the development of an autonomous multi-robot search and rescue assistance platform capable of sensor fusion and object detection in embedded devices using novel lightweight AI models.

\subsection{Significance}

Apart from the computational constraints, the specific application scenario poses unique challenges, since we are not only interested in analyzing the sensory data from the UAVs and the boat, but also in enabling active interaction among these vehicles to make the search process more efficient. For example, if a UAV sees something that looks like a human it should take appropriate actions to go toward the area of interest or request assistance of another UAV for cooperative examination (active vision and tracking). Furthermore, UAVs may have simultaneous sightings of different objects, which need to be ranked by their reliability and urgency. Therefore, methods for intelligent UAV vision and perception will be developed toward this end. These will be combined with distributed formation control algorithms that will allow UAVs to reconfigure their spatial distribution as a function of sightings and their ranking, dynamic needs in communication range and bandwidth, active vision and tracking, and the density of UAVs required in different areas. There might be a high concentration of objects in some areas, and more UAVs be required there to fly closer to the surface and classify those objects, while others explore larger areas from higher above the sea to perform a preliminary reconnaissance. All these decisions must be performed autonomously, either by multiple UAVs independently or in cooperation with the central hub installed on the rescue vessel.

In this document, we use the following terminology. A \textit{multi-UAV system} refers to a group of UAVs cooperatively performing a given task. A \textit{heterogeneous multi-robot platform} (or just Platform hereinafter) refers to the combination of a vessel and a multi-UAV system with the technology developed for object recognition and classification, communication, and coordination frameworks.

\subsection{Research Questions}

The AutoSOS project develops an autonomous Platform to support maritime SAR operations that relays on an autonomous vessel as a central hub for computation and communication. The project concentrates on developing algorithms for perception and control in a heterogeneous multi-robot system that enables autonomous SAR operations. carried out by collaborative UAVs and a vessel - a heterogeneous multi-robot system. To this end, we will explore the following two fundamental topics:

\begin{itemize}
    \item Autonomous coordination of UAVs from the perspective of multi-agent systems to maximize area coverage, and ensure connectivity of all UAVs with the central hub with multi-hops, if necessary. At the same time, formation control algorithms need to take into account not only the changing communication topology within the UAV network but also a ranking of sightings and distribution of computational tasks. The multi-UAV system will be able to autonomously and dynamically decide on both its objective spatial configuration and task allocation, based on novel multi-agent and multi-objective optimization methods.
    \item Design, development, and implementation of robust, adaptive, and lightweight multi-modal object detection models that will run on small embedded devices on-board the UAVs. Optimization of deep learning methods for specific tasks. This includes algorithms for autonomous decision-making at the edge devices (UAVs) level, and autonomous task migration and communication priority decisions at the multi-UAV system level.
\end{itemize}

\subsection{Contribution and Structure}

We present a multidisciplinary approach heavily influenced by a wide variety of cutting-edge technologies to develop new methods for autonomous cooperation and execution within a multi-robot platform for SAR operations. Research in AutoSOS will use as a basis the latest techniques in adaptive machine learning models for embedded (edge) devices, theory of multi-agent systems in consensus and formation control, and the world of Internet of Things (IoT). From the IoT, we take the state-of-the-art developments in distributed computation using edge computing and integrate it into the control flow of a heterogeneous multi-robot system. This provides a robust and fault-tolerant framework for communication and control. However, the main limiting factor of many applications for multi-agent systems is situational awareness in unknown environments. That is why we will develop novel adaptive and lightweight multi-modal machine learning models, in combination with the computational load distribution inherent to an edge computing structure.

Formation control of a multi-drone system has been previously studied independently of the problems of enhancing a ground or surface unit's situational awareness with aerial vehicles, the communication layer, and computational task distribution, and also separately from the integration of adaptive multi-modal deep learning techniques for advanced autonomous vision. The main impact of this project will be on developing new methods and algorithms that are able to integrate into a single multi-agent control loop adaptive deep learning techniques for advanced vision, communication constraints, spatial awareness, and computation distribution. By jointly optimizing, and further developing the apparently independent \textit{things} (methods and technologies) we will achieve a more efficient autonomous \textit{everything.}

In summary, this paper presents the main objectives and research questions behind the AutoSOS project, together with a comparison with the state of the art in multi-UAV systems for search and rescue operations. We believe this conceptual overview will help to push towards more intelligent multi-UAV systems, and that the research directions described in this paper will benefit other researchers to better understand the current efforts being put in this field.

The remainder of this paper is organized as follows. Section 2 introduces related projects and other previous works. In Section 3, we describe the proposed architecture and justify the needs for tighter integration of active perception and multi-agent multi-objective path planning algorithms. Then, Section 4 presents potential application scenarios. Finally, Section 5 concludes the work.
\begin{table*}[ht!]
    \centering
    \caption{Recent EU projects in SAR Robotics}
    \label{tab:projects}
    \begin{tabular}{@{}lccccccccc@{}}
        \toprule
        & Multi-robot & Autonomous & Autonomous & Autonomous & Multi-UAV & Heterogeneous & On-board & Mesh & Distributed \\
        & system & UGV & USV & UAV & system & robots & Processing & Network & control \\
        \midrule
        \textbf{NIFTi} & \cmark & \cmark & - & \cmark & - & \cmark & - & - & - \\
        \textbf{ICARUS} & \cmark & \cmark & \cmark & \cmark & \cmark & \cmark & - & - & - \\
        \textbf{TRADR} & \cmark & \cmark & - & \cmark & \cmark & \cmark & - & - & - \\
        \textbf{SmokeBot} & - & \cmark & - &  & - & - & \cmark & - & - \\
        \textbf{CENTAURO} & \cmark & - & - & - & - & - & - & - & - \\
        \textbf{AutoSOS} & \cmark & - & \cmark & \cmark & \cmark & \cmark & \cmark & \cmark & \cmark \\
        \bottomrule
    \end{tabular}
\end{table*}

\section{Related Projects}
\label{sec:related}

Most previous research efforts in the area of SAR robotics have mainly focused on the design of individual robots autonomously operating in emergency scenarios, such as those presented in the European Robotics League Emergency Tournament. The previous works in multi-UAV and multi-robot systems for SAR either use an external control center for route planning and monitoring \cite{cooperative_multi_robot_surveillance}, rely on a static base station and predefined patterns for finding objectives \cite{scherer2015autonomous}, or have predefined interaction between different units \cite{multi_robot_maritime}. 

Recent international EU projects designing and developing autonomous multi-robot systems for search and rescue operations include the NIFTi project (natural human-robot cooperation in dynamic environments)~\cite{kruijff2014designing}, ICARUS (unmanned search and rescue)~\cite{cubber2017search}, TRADR (long-term human-robot teaming for disaster response)~\cite{de2018persistent, gawel2018x, freda20193d}, SmokeBot (mobile robots with novel environmental sensors for inspection of disaster sites with low visibility)~\cite{fritsche2016radar, wei2016multi}. Other projects, such as CENTAURO (robust mobility and dexterous manipulation in disaster response by full body telepresence in a centaur-like robot), have focused on the development of more advanced robots that are not fully autonomous but controlled in real-time~\cite{klamt2019flexible}. A comparison of these projects and AutoSOS is provided in Table~\ref{tab:projects} from a system-level perspective. We list the types of robots utilized in the different projects: unmanned ground vehicles (UGVs), unmanned surface vehicles (USVs), and unmanned aerial vehicles (UAVs).

In NIFTi, both UGVs and UAVs were utilized for autonomous navigation and mapping in harsh environments~\cite{kruijff2014designing}. The focus of the project was mostly on human-robot interaction and on distributing information for human operators at different layers. Similarly, in the TRADR project, the focus was on collaborative efforts towards disaster response of both humans and robots~\cite{de2018persistent}, as well as on multi-robot planning~\cite{gawel2018x, freda20193d}. In particular, the results of TRARD include a framework for the integration of UAVs in search and rescue missions, from path planning to a global 3D point cloud generator~\cite{surmann2019integration}. The project continued with the foundation of the German Rescue Robotics Center at Fraunhofer FKIE, where broader research is conducted, for example, in maritime search and rescue~\cite{guldenring2019heterogeneous}. In ICARUS, project researchers developed an unmanned maritime capsule, a large UGV, and a group of UAVs for rapid deployment, as well as mapping tools, middleware software for tactical communications, and a multi-domain robot command and control station~\cite{cubber2017search}. While these projects focused on the logarithmic aspects of search and rescue operation, and on the design of multi-robot systems, in Smokebot the focus was on developing sensors and sensor fusion methods for harsh environments~\cite{fritsche2016radar, wei2016multi}.

\section{AutoSOS Platform}

\begin{figure*}[ht!]
    \centering
	\includegraphics[width=0.75\textwidth]{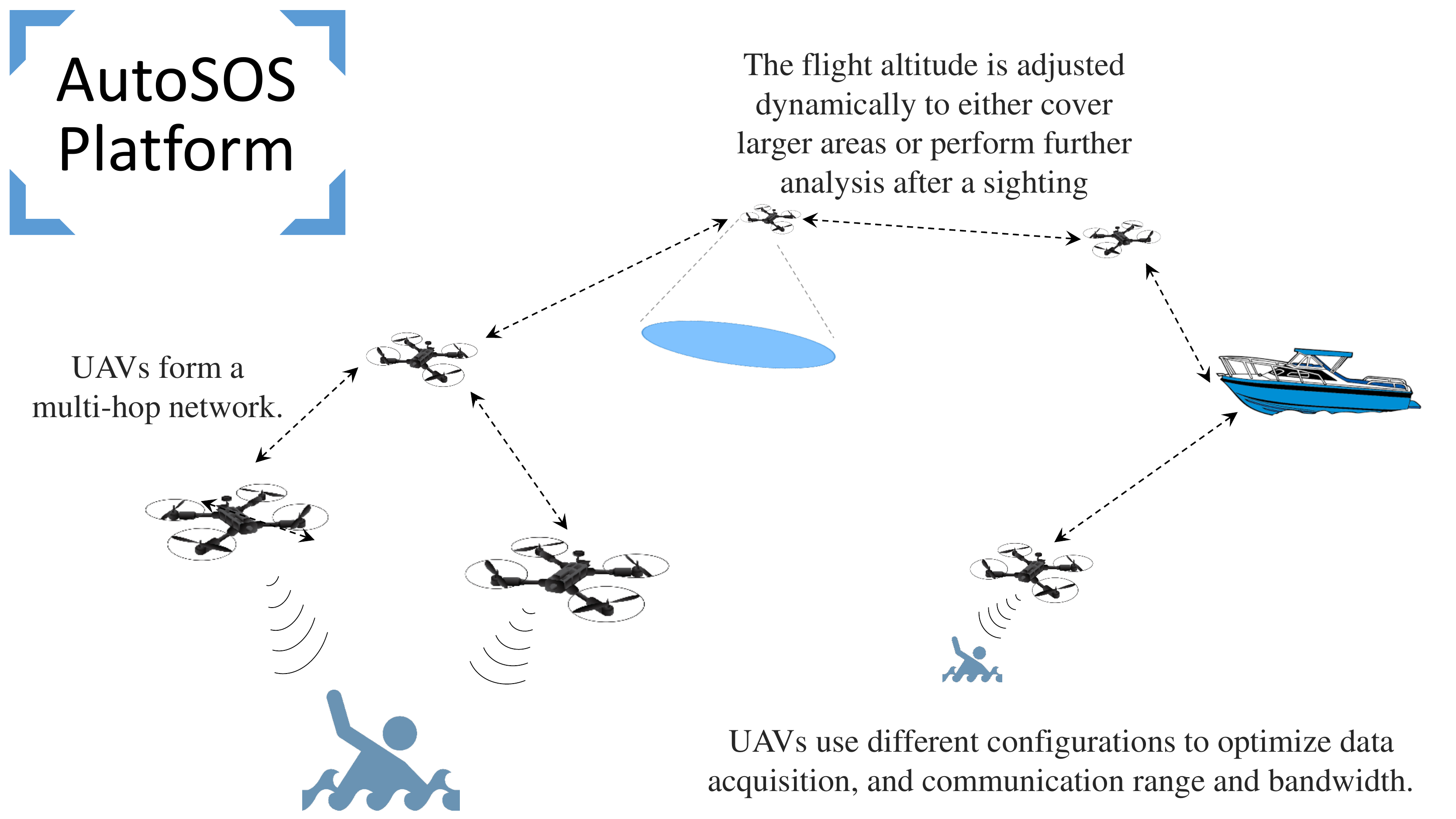}
    \caption{AutoSOS Platform architecture.}
    \label{fig:platform}
\end{figure*}

\begin{figure}[t]
    \centering
	\includegraphics[width=0.48\textwidth]{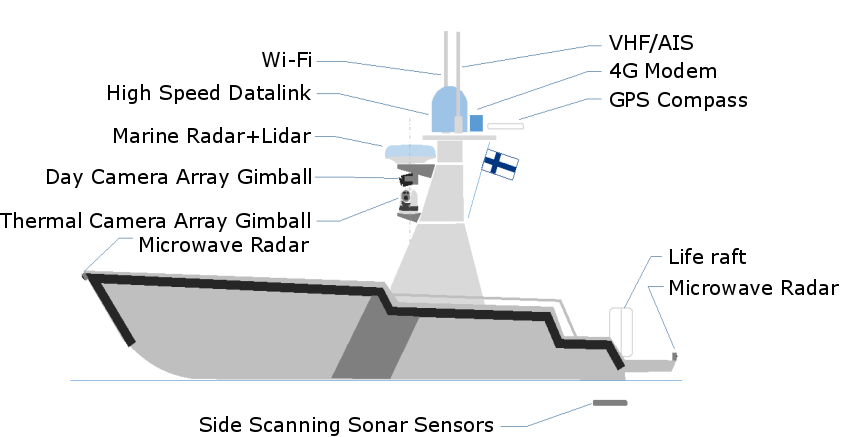}
    \caption{Illustration of the rescue vessel.}
    \label{fig:boat}
\end{figure}

AutoSOS proposes the design and development of a heterogeneous platform of autonomous vehicles consisting of a rescue vessel and a multi-UAV system illustrated in Figure\:\ref{fig:platform}. In this project, the rescue boat is a research vessel previously used in the aCOLOR project (TAU, Alamarin Jet)~\cite{villa2020acolor} shown in Figure\:\ref{fig:boat}. The boat is to be autonomous, but here we consider only the autonomy of the search operation from the algorithmic point of view. The other aspects of the boat autonomy (such as steering and docking) are beyond the scope of the AutoSOS project. For the rescue operation, the boat is assumed to be manned with rescue personnel and equipped with different sensors, such as RGB cameras, thermal cameras, marine radar, and LiDAR. All these sensors can be exploited to detect and recognize objects and persons and the gathered information will be combined with sensor data from UAVs using novel multi-modal techniques.

The UAVs are deployed from the rescue vessel and given an objective search area. From that point, they operate in a cooperative manner as an autonomous multi-agent system. The vessel is used as a central hub for computational offloading and communication base. Initially, UAVs plan their paths in order to maximize area coverage flying at a high altitude that enables object detection but not classification. Advanced lightweight and adaptive deep learning algorithms enable a preliminary assessment of the area in terms of object/person detection and classification. Then, active vision and tracking techniques are used to analyze closer the objects that have been found. A ranking system allows concurrent sightings to be classified with a given priority, and this is then used as feedback in the formation control algorithms, with UAVs more likely moving toward areas with a higher density of sightings. At the same time, another parameter that is used as feedback for formation control is the distance between the vessel and the sightings and the spatial distribution of UAVs. If needed, the UAVs will reconfigure their positions to ensure a robust communication link between the vessel and the UAVs. Finally, the bandwidth of the communication link will also be taken into account when more accurate data processing is required: as UAVs have limited computational capabilities, they might need to transmit high-resolution images or other data to the boat for \textit{deeper} processing and analysis.

\subsection{Distributed Formation Control and Task Allocation}

The first step is the design and development of methods for autonomous distributed task allocation and formation reconfiguration. This covers the design of a cooperation framework for the vessel and the multi-UAV system. The framework includes methods to distribute the computational load, strategies for active vision and active tracking in UAVs, design of fallback communication options, and path planning for efficient search in the objective area. All this wraps up with the development of formation control algorithms able to maintain robust communication links, maximize area coverage, and optimize the use of computational resources.

We will use as a starting point for formation control algorithms and consensus methods from the theory of multi-agent systems, and methods for collaborative sensing and cooperatively enhancing situational awareness \cite{queralta2020enhancing, queralta2020blockchain}. This will then be combined with a set of at least two different communication technologies (short-range, high bandwidth and long-range, decreasing bandwidth). For these,  a relatively recent solution ultra-wideband (UWB) wireless communication~\cite{shule2020uwbbased}. UWB-based localization systems have been applied to multi-robot and multi-UAV systems for accurate localization~\cite{queralta2020uwb}. Moreover, UWB has the potential for simultaneous communication and localization. We will continue to work on extending our current UWB-based solutions, including mobile localization systems capable of centimeter-level accuracy~\cite{almansa2020autocalibration}
In formation control algorithms, an external input to the system is the configuration requirements in terms of spatial distribution~\cite{mccord2019progressive}. In AutoSOS, agents will autonomously arrange themselves to optimize area coverage initially using distributed pattern configuration algorithms that we will design extending our previous work~\cite{queralta2019indexfree, queralta2019progressive}. As the search mission proceeds, new parameters will be taken into account such as active vision and tracking of findings. A ranking system will be implemented where the output of machine learning models will be used as feedback in the formation control process. Moreover, UAVs with less priority will rearrange to ensure a stable multi-hop communication link between the vessel and the UAVs with highest ranked findings. Finally, the communication range and required bandwidth if sensor data needs to be transmitted to the vessel will also be taken into account in the rearrangement process, dynamically switching and optimally exploiting the different communication technologies available. To this end, machine learning algorithms will be employed and combined with  intelligent control methods to improve the efficiency and accuracy of the information processing~\cite{active_perception}, balancing the load between on-board computation at the edge (UAVs) and offloading computation to the vessel, and taking into account the quality of communication channels~\cite{edgeAILora, qingqing2019odometry, qingqing2019fpga}. If possible, we will also contemplate the possibility of distributing computation among UAVs, so that a group of near UAVs can be divided between sensing UAVs and processing UAVs, the former ones in charge of obtaining data and distributing it to the latter ones, where different frames are analyzed and results shared within the system.

\subsection{Lightweight Edge AI}

The second key area is the design and development of algorithms that enable efficient and accurate analysis of data from multiples sensors at the edge (UAVs) and on the boat and development of adaptive lightweight deep learning algorithms that optimize the use of power and computational resources. Several data fusion methods and multimodal algorithms will be investigated to further increase the accuracy and reliability of the perception algorithms. AutoSOS will focus on a) intra-device multi-modal fusion and b) inter-device multi-modal fusion. Intra-device multi-modal fusion refers to fusing the information from various sensors of on the same device, e.g., a UAV. For example, we will consider fusing the information collected from the visual camera and thermal cameras to increase the object detection and segmentation accuracy. On the other hand, inter-device multi-modal fusion refers to fusing information collected from sensors that exist on different devices to improve the performance of the employed algorithms. Therefore, AutoSOS will consider fusing point-cloud data collected from the lidar on the boat with the information dynamically gathered from the UAVs. Moreover, methods for fusing and ranking simultaneous detections, from different UAVs and multiple views, will be also considered to guide the search strategies (as described in the previous subsection). The research on data fusion will continue the earlier ongoing research of the consortia \cite{Gabbouj_multiview1, Gabbouj_multiview2, Gabbouj_multiview3, Gabbouj_multiview4} and use data and experience we have previously collected in the aColor project \cite{villa2020acolor, taipalmaa2019watersegmentation}. 
    
Furthermore, even though existing state-of-the-art deep learning methods are effective, they require a lot of computational resources, which is problematic on the environment where we expect quick processing with a low energy consumption on a lightweight hardware. Therefore, we will consider neural network distillation \cite{distillation} and knowledge transfer methods \cite{knowledge_transfer} to transfer the knowledge from larger and computationally heavy models into smaller and faster ones, ensuring that the proposed methods will be able to run in real-time on the available hardware. Methods that dynamically adapt the inference graph of the models to the available computational resources, such as \cite{adaptive_graphs}, will be adapted to the needs of the current application to further improve the speed of the models and ensure the optimal utilization of the hardware. This will allow for having higher quality prediction when the computational load is low, while still ensuring the real-time operation even when the UAVs have to process a great amount of information. Recent studies by the consortia \cite{Passalis_adaptive, passalis2020exits} showed that this is a promising approach for low-power embedded devices. Our previous work on energy-efficient hardware accelerators for deep learning will also be leveraged in this case~\cite{cgra_nn, low_power_cnns}, as well as embedded AI for edge computing~\cite{metwaly2019thermal}.
\section{Application Scenarios}

Research outcomes from AutoSOS will directly contribute towards next-generation autonomous rescue operations, which have a big potential in saving human lives. Researchers behind AutoSOS have been in contact with police forces and rescue teams in Finland and abroad, who acknowledged maritime SAR operations as a very clear use case of the technology that will be developed within the project. In particular, AutoSOS will carry out field tests and experiments in the Mediterranean Sea together with the Alicante Search and Rescue Group.

Even though search tasks at sea are often thought to be related to harsh weather conditions and adverse environments, this is not always the case. In touristic areas that are typical beach holiday destinations, maritime SAR operations happen with increasing frequency in the coast. This happens even during daytime on normal summer days with good environmental conditions, when UAVs can be deployed at their full potential. Therefore, the AutoSOS Platform can have a direct impact in these operations and enhance the public service toward more efficient SAR missions.

Besides rescue operations, the same methods can be in the future applied in different search tasks such as environmental monitoring, and in different environments outside out the sea. Even though many of the perception and vision algorithms that use deep learning will be trained for maritime scenarios, the same platform architecture and equipment can be employed for search operations in mountains, fields or large areas in general. 

\section{Conclusion and Future Work}

We have presented the main goals of AutoSOS, a project where we design and develop a multi-UAV system for assisting in maritime search and rescue operations. The key research areas in this project are distributed coordination algorithms for multi-UAV systems and lightweight computer vision algorithms able to run on embedded platforms in real time. Both directions will come together with the design and development of distributed multi-agent active perception and active tracking algorithms running at the edge.

\section*{Acknowledgements}

This work was supported by the Academy of Finland's AutoSOS project with grant number 328755.

\bibliographystyle{IEEEtran}
\bibliography{main.bib}

\end{document}